\documentclass{article} 
\usepackage[final]{colm2026_conference}

\usepackage{microtype}
\usepackage{hyperref}
\usepackage{url}
\usepackage{amsmath}
\usepackage[capitalize,noabbrev]{cleveref}

\usepackage{amssymb}
\usepackage{mathtools}
\usepackage{amsthm}

\usepackage{graphicx}
\usepackage{subcaption}
\usepackage{wrapfig}
\usepackage{booktabs}
\usepackage{multirow}
\usepackage{makecell}
\usepackage{multicol}

\usepackage{pifont}
\usepackage[misc]{ifsym}

\usepackage[table]{xcolor}
\definecolor{mygreen}{RGB}{0,120,90}
\definecolor{lightpurple}{RGB}{244,232,255}
\definecolor{maskpurple}{RGB}{123,104,214}
\definecolor{reversemagenta}{RGB}{190,65,125}
\definecolor{deepblue}{HTML}{3B6FB6}

\definecolor{darkblue}{rgb}{0, 0, 0.5}
\hypersetup{colorlinks=true, citecolor=mid, linkcolor=primary, urlcolor=mid}

\renewcommand{\cite}[1]{\citep{#1}}

\title{Trust Region On-Policy Distillation}

\author{
\mbox{
Xingrun Xing$^{1}$, Haoqing Wang$^{1}$, Boyan Gao$^{2}$, Ziheng Li$^{1, 3}$, and Yehui Tang$^{1{~\textrm{\Letter}}}$ }\\
$^1$ Samsung Research, Beijing, China \\
$^2$ University of Oxford $^3$ Peking University \quad\quad \\
\texttt{xingrun.xing}@partner.samsung.com\quad\texttt{yehui.tang}@samsung.com \\
$^\textrm{\Letter}$~Corresponding Author
}

\begin{document}

\maketitle
\begin{abstract}
On-Policy Distillation (OPD) is a fundamental technique for efficient post-training of large language models (LLMs), with broad applications in agent learning, multi-task enhancement, and model compression. However, OPD training becomes unstable when the teacher and student distributions differ substantially, as teacher supervision on student-generated tokens may yield unreliable policy gradients and even cause optimization failure.
This work addresses reliable on-policy token-level supervision through credit assignment strategies, and proposes Trust Region On-Policy Distillation, TrOPD. 
It features the following characteristics:
\textbf{1) Trust-Region On-Policy Learning:} TrOPD performs OPD only in regions where the teacher provides reliable supervision, mitigating the optimization difficulty of the $K_1$ reverse-KL estimator under distribution mismatch. 
\textbf{2) Outlier Estimation:} For outlier regions, we explore gradient clipping, masking, and forward-KL estimation to reduce the adverse effects of unreliable supervision. 
\textbf{3) Off-Policy Guidance:} The student continues generation from teacher prefixes and uses forward KL to imitate off-policy guidance, encouraging on-policy exploration toward reliable regions.
Experiments show that TrOPD consistently outperforms SoTA OPD baselines, including OPD, EOPD, and REOPOLD, across mathematical reasoning, code generation, and general-domain benchmarks. The project homepage is available at \href{https://github.com/Xingrun-Xing2/TrOPD/tree/main}{GitHub}.
\end{abstract}

\section{Introduction}

\begin{figure}[h!]
\centering
\includegraphics[width=\columnwidth]{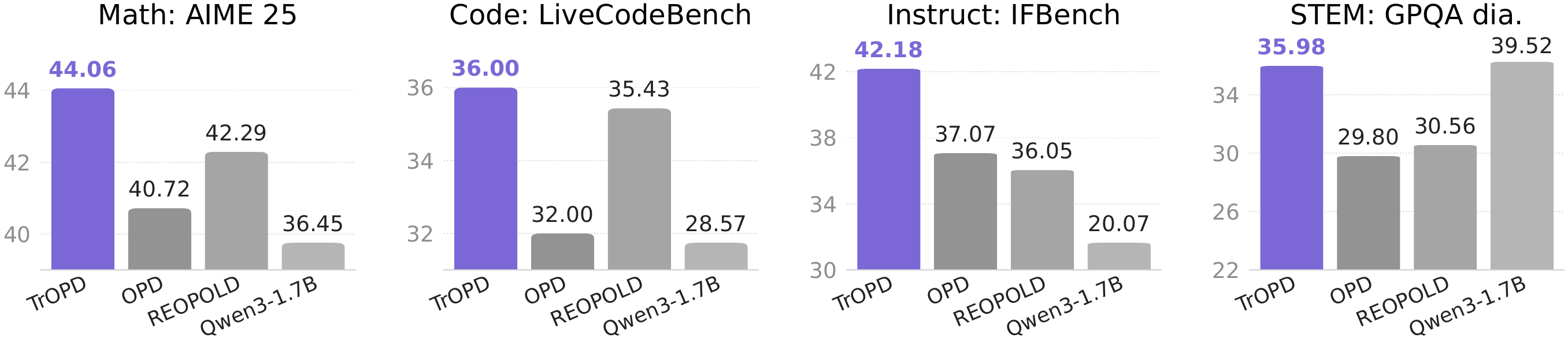}
\vspace{-0.1in}
\caption{
    Performance comparison of TrOPD and baselines. The OPD methods, including TrOPD, OPD, and REOPOLD \cite{ko2026scaling}, are trained on Qwen3-SFT-1.7B, which is finetuned from Qwen3-1.7B-Base via supervised finetuning.
}
\label{f1}
\vspace{-0.05in}
\end{figure}

Recent Large Reasoning Models (LRMs) \cite{zhang2025100,chen2024huatuogpto1medicalcomplexreasoning} improve performance by scaling test-time reasoning and have achieved expert-level performance in mathematics~\cite{ren2025deepseek}, code generation~\cite{anthropic2025claude}, and agent tasks~\cite{ghareeb_robin_2025}.
However, their substantial inference costs motivate the development of Small Reasoning Models (SRMs)~\cite{zhao2025mobilellm} for resource-efficient deployment.
Conventional off-policy distillation~\cite{kim2016sequence} trains students to imitate outputs generated by strong teacher models.
Since training relies on teacher-generated trajectories while inference follows student-generated ones, this paradigm suffers from exposure bias, especially in long-chain-of-thought reasoning.
On-Policy Distillation (OPD)~\cite{lu2025onpolicy,agarwal2024onpolicy} mitigates this issue by training directly on student-generated trajectories, making it an efficient approach for SRMs.

Despite their potential efficiency advantages, existing OPD methods often suffer from training instability due to unreliable supervision. 
When the teacher and student distributions diverge substantially, student-generated trajectories may fall outside the teacher's reliable supervision region, yielding erroneous policy gradients and potentially causing training collapse.
Moreover, reasoning-oriented OPD cannot afford full-vocabulary supervision due to its prohibitive memory overhead~\cite{agarwal2024onpolicy}. 
It therefore typically relies on KL divergence estimators~\cite{lu2025onpolicy}, which may further reduce the reliability of the supervision signal.

However, reliable OPD for reasoning tasks remains non-trivial. 
This work establishes a unified benchmark to systematically study this challenge from three perspectives: 
(1) multi-domain evaluation, covering mathematics, code generation, and STEM reasoning; 
(2) diverse OPD strategies, comparing representative conventional and recent methods under unified training settings; and 
(3) memory-efficient KL estimation, implementing the $K_1$ and top-$k$ estimators to enable long-response distillation under practical memory constraints. 
The resulting evaluation reveals that existing methods fail to effectively suppress erroneous policy gradients. 
Furthermore, naive reward clipping, as adopted by REOPOLD~\cite{ko2026scaling}, may remove informative supervision together with outlier gradients, resulting in a performance bottleneck.

\begin{figure*}[t]
\centering
\includegraphics[width=\textwidth]{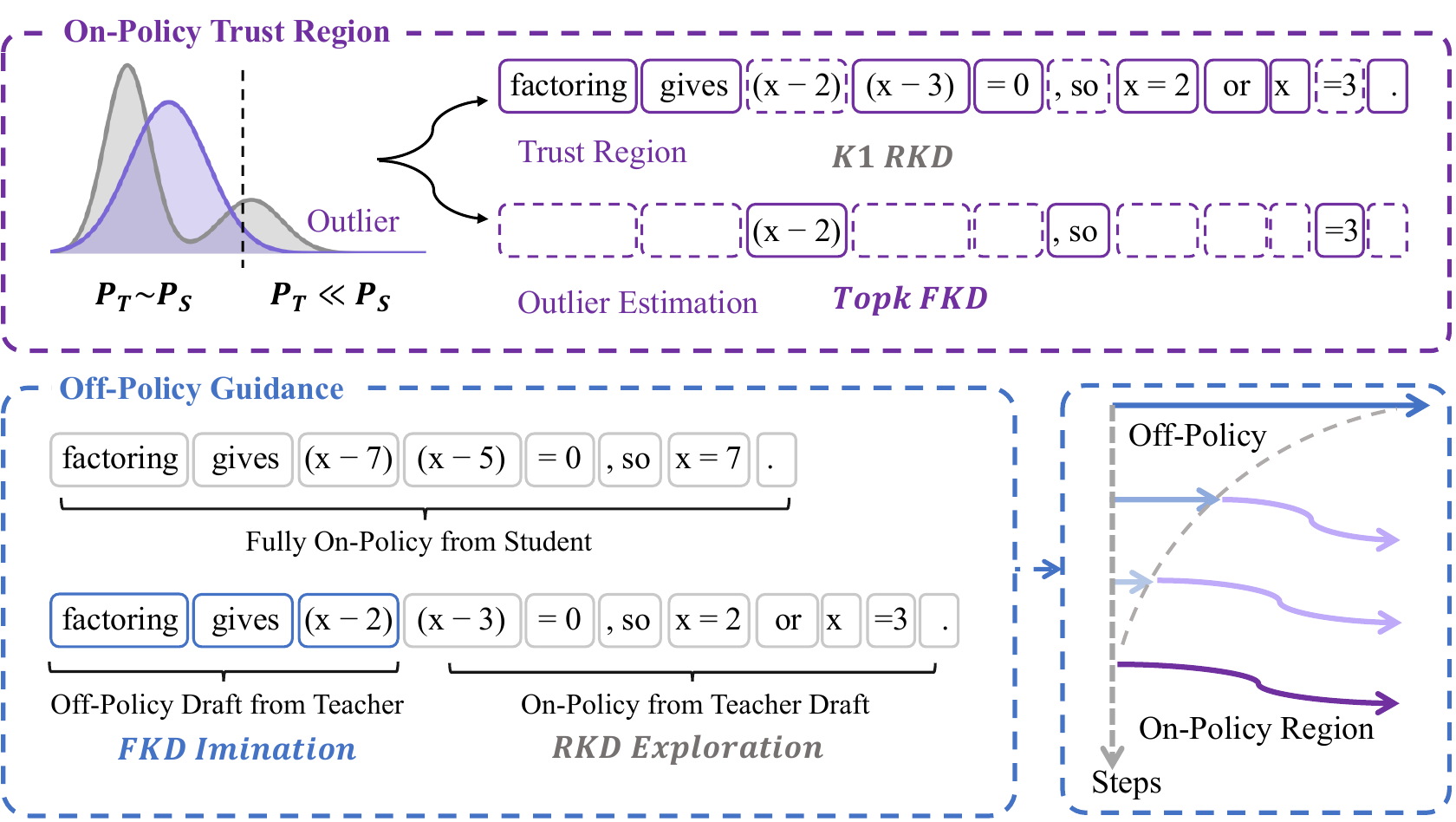}
\vspace{-0.1in}
\caption{
    Overview of Trust Region On-Policy Distillation. 
    For the on-policy component, student-generated tokens are divided into the trust region and outliers.
    The student model is further guided by teacher-generated responses.
}
\label{f2}
\vspace{-0.1in}
\end{figure*}

To improve the reliability of teacher supervision, this work proposes Trust Region On-Policy Distillation (TrOPD), which partitions student-generated tokens according to their supervision reliability. 
As shown in Figure~\ref{f2}, TrOPD determines whether a token falls into a teacher-verifiable trust region according to the decoding agreement ratio between the teacher and student models. For outliers, we employ a top-$k$ forward-KL estimator to preserve informative reward signals while avoiding unreliable policy gradients. To further encourage the student to generate within teacher-verifiable regions, we introduce off-policy guidance, which performs imitation learning from teacher-generated trajectories. As shown in Figure~\ref{f1}, TrOPD substantially improves OPD by $+3.34$, $+4.00$, $+5.11$, and $+6.18$ points on math, code, instruction following, and STEM benchmarks, respectively.

Our contributions are summarized as follows:
\begin{itemize}
    \item We establish a general benchmark for reasoning-oriented OPD and identify the supervision reliability issue in OPD.
    \item We propose Trust Region On-Policy Distillation (TrOPD), achieving high-quality and stable reasoning optimization.
    \item We train small reasoning models based on TrOPD, further advancing the reasoning capabilities of small language models.
\end{itemize}

\section{Related Works}

\paragraph{Reasoning Language Models.}
Reasoning ability \cite{zhang2025100,chen2024huatuogpto1medicalcomplexreasoning} has become a major driver of performance improvements in large language models (LLMs), initially elicited through reasoning prompts. More recently, reasoning capabilities have been acquired through reinforcement learning \cite{zhang2025100}; supervised finetuning \cite{kimiteam2025kimik2openagentic}; and on-policy distillation. Reasoning is also increasingly integrated with other core capabilities of LLMs, such as agentic and multimodal \cite{kimi-researcher2025,bai2025qwen3} abilities. However, how to acquire strong reasoning capabilities for SRMs remains underexplored.

\paragraph{Knowledge Distillation.}
Knowledge distillation was originally introduced by Hinton et al. \cite{hinton2015distilling} to efficiently train compact models. In recent years, distillation for generative language models has primarily relied on sequence-level knowledge distillation \cite{kim2016sequence}, an off-policy approach that performs supervised fine-tuning on teacher-generated responses. More recently, full-vocabulary on-policy distillation methods \cite{ko2024distillm,ko2025distillm,xu2025speculative}, such as MiniLLM \cite{gu2024minillm} and GKD \cite{agarwal2024onpolicy}, have been developed to mitigate the exposure bias \cite{agarwal2024onpolicy}. For reasoning models, KL objectives based on the $k_1$ estimator \cite{lu2025onpolicy,yang2026learning} have been applied to effectively improve the reasoning performance in post-training stage.

\section{Problem Formulation}

\subsection{Distillation for Language Models}
Different from off-policy distillation trained from the teacher of generations (ToGs), 
On-policy distillation instead trains on student of generations (SoGs) to mitigate exposure bias. In reverse KL divergence (RKL)-based OPD, the sequence-level objective can be written as $D_{\mathrm{KL}}(\pi_S \| \pi_T) = \mathbb{E}_{x \sim \pi_S}\left[\log \frac{\pi_S(x)}{\pi_T(x)}\right]$, whose gradient naturally takes a policy-gradient form \cite{gu2024minillm}: the student samples trajectories from its own policy and is rewarded for generating sequences assigned high probability by the teacher. Since the expectation is taken over the student distribution, RKL strongly penalizes student outputs that fall into low-probability regions of the teacher, while imposing little direct penalty on teacher modes not explored by the student, thereby exhibiting mode-seeking behavior. 
However, conventional OPD computes RKL or Jensen--Shannon divergence (JSD) over the full vocabulary, making memory overhead a significant bottleneck for long generation tasks.

\begin{table*}[t]
\centering
\small
\setlength{\tabcolsep}{4.5pt}
\renewcommand{\arraystretch}{1.12}
\resizebox{\textwidth}{!}{%
\begin{tabular}{lcccccc}
\toprule
Method & FKL Objective & RKL Objective & AIME 24 & AIME 25 & AMC 23 & Avg. \\
\midrule
DeepSeek-Qwen2.5-1.5B 
& -- 
& -- 
& 28.64 & 24.16 & 71.01 & 41.27 \\

OPD (RKL) 
& -- 
& $\log \pi_{T}/\pi_{S}$ 
& 35.83 & 29.16 & 75.39 & 46.79 \\
\midrule

\multicolumn{7}{l}{\emph{\quad \textbf{Distribution Mixing Strategies}}} \\

FKL 
& $\sum_{v \in \mathcal{V}^{T}_{k}} \pi_{T,v} \log (\pi_{S,v}/\pi_{T,v})$ 
& -- 
& 0.00 & 0.00 & 4.21 & 1.40 \\

JSD 
& $\beta \sum_{v \in \mathcal{V}^{T}_{k}} \pi_{T,v} \log (\pi_{M,v}/\pi_{T,v})$ 
& $(1-\beta) \log (\pi_{M}/\pi_{S})$ 
& 37.91 & 30.72 & 75.07 & 47.90 \\
\midrule

\multicolumn{7}{l}{\emph{\quad \textbf{Entropy-Aware Token Selection Strategies}}} \\

Entropy OPD 20\% 
& -- 
& $\mathbb{I}[\mathrm{H}>\tau]\log \pi_{T}/\pi_{S}$ 
& 35.52 & 29.06 & 73.82 & 46.13 \\
\midrule

\multicolumn{7}{l}{\emph{\quad \textbf{Outlier-Aware Token Selection Strategies}}} \\

Clip Outlier 
& -- 
& $\max(\log \pi_{T}/\pi_{S},\tau)$ 
& 36.97 & 30.83 & 75.78 & 47.86 \\

\rowcolor{lightpurple}
Mask Outlier 
& -- 
& $\mathbb{I}[\mathrm{R}>\tau] \log \pi_{T}/\pi_{S}$ 
& 37.08 & 30.62 & 75.46 & 47.72 \\

\rowcolor{lightpurple}
FKL Outlier 
& $\overline{\mathbb{M}} \sum_{v \in \mathcal{V}^{T}_{k}} \pi_{T,v} \log (\pi_{S,v}/\pi_{T,v})$ 
& $\mathbb{M} \log \pi_{T}/\pi_{S}$ 
& 39.16 & 29.89 & 77.96 & 49.00 \\

\rowcolor{lightpurple}
TrOPD 
& $\overline{\mathbb{M}} \sum_{v \in \mathcal{V}^{T}_{k}} \pi_{T,v} \log (\pi_{S,v}/\pi_{T,v})$ 
& $\mathbb{M} \log \pi_{T}/\pi_{S}$ 
& 38.54 & 32.50 & 78.51 & \textbf{49.85} \\
\bottomrule
\end{tabular}}
\caption{
Comparison of OPD methods on math-domain reasoning benchmarks.
All OPD methods are trained with Skywork-OR1-Math-7B as the teacher model.
}
\label{t1}
\vskip -0.1in 
\end{table*}

\subsection{OPD for Reasoning Models}

Recent on-policy distillation (OPD) methods employ token-level reverse KL as the reward and optimize it using policy gradient.
However, computing reverse KL over the full vocabulary incurs 
$\mathcal{O}(n \cdot k)$ memory overhead, where $n$ is the sequence length 
and $k$ is the vocabulary size. 
Traditional instruction LLM distillation methods, such as GKD and speculative KD, compute KL divergence over the full vocabulary, achieving stable optimization: 
\begin{equation}
\mathcal{J}^{\mathrm{KD}}
=
-\mathrm{RKL}(\pi_S \parallel \pi_T)
=
-\sum_{x \in \mathcal{V}}
\pi_S
\log \frac{\pi_S}{\pi_T}.
\end{equation}

In contrast, recent reasoning-oriented models scale performance by extending the reasoning length, which significantly increases output sequence length and makes memory consumption a major bottleneck for model distillation. To address this issue, Thinking Machine Lab proposes using the $K_1$ estimator to obtain an unbiased estimate of the KL divergence, leading to the following optimization objective:
\begin{equation}
\mathcal{J}^{\mathrm{KD}}
=
-\mathrm{RKL}(\pi_S \parallel \pi_T)
=
-\mathbb{E}_{x \sim \pi_S}
\left[
\log \frac{\pi_S}{\pi_T}
\right].
\end{equation}
However, the $K_1$ estimator suffers from two key optimization bottlenecks:

\paragraph{Significant policy-gradient outliers.}
When the discrepancy between the teacher and student distributions is large, the teacher may assign extremely low probabilities to trajectories sampled from the student policy, i.e., $\mathbb{E}_{x \sim P_S}\left[\pi_T(x)\right] \approx 0$. In such low-confidence regions, the $K_1$-based policy-gradient signal can become extremely negative, i.e., $\nabla \mathcal{J} = \frac{1}{\pi_S(x)} \log \frac{\pi_T(x)}{\pi_S(x)} \rightarrow -\infty$. Therefore, student-generated trajectories that receive extremely low confidence from the teacher induce significant policy-gradient outliers, which destabilize OPD optimization and limit its potential final performance.

\paragraph{Low-quality student of generation (SoG).}
Since OPD is optimized exclusively on trajectories sampled from the student policy, the student may struggle to generate high-quality responses for challenging problems. As a result, low-quality SoG trajectories restrict the effective optimization space and prevent the student from receiving informative supervision from higher-quality responses.

\section{Trust Region Distillation}

\subsection{Benchmarking OPD Baselines}

Compared with conventional OPD based on full-vocabulary distributions, OPD for long-thinking reasoning models remains an emerging research direction. Existing studies are often conducted under different experimental configurations, making it difficult to directly compare recent SoTA methods. Therefore, there is an urgent requirement to benchmark recent OPD methods under a unified setting. In this section, we first evaluate representative OPD methods by focusing on two fundamental questions: 
(1) Do divergence objectives developed for full-vocabulary OPD remain effective for recent token-level OPD based on the $K_1$ estimator? 
(2) Under a unified experimental setting, how do existing advanced methods compare fairly in terms of performance and generalization?

\paragraph{Divergence Evaluation.}
Due to the asymmetry of KL divergence, previous full-vocabulary distillation methods typically adopt Forward KL (FKL) for mode covering and Reverse KL (RKL) for mode seeking. More generally, GKD \cite{agarwal2024onpolicy} employs JSD to balance FKL and RKL. Under the memory constraints of long-thinking distillation, we implement FKL over the top-$k$ tokens of the teacher distribution and implement RKL using the token-level $K_1$ estimator. Specifically, the top-$k$ FKL objective is defined as:
\begin{equation}
    \mathcal{J}_{\mathrm{FKL}}^{\mathrm{top}\text{-}k}
    =
    -\sum_{v \in \mathcal{V}^{T}_{k}}
    \pi_{T,v}
    \log
    \frac{\pi_{T,v}}{\pi_{S,v}} .
\end{equation}
Given the definitions of top-$k$ FKL and $K_1$-based RKL, 
the generalized JSD objective with a balancing coefficient $\beta$ 
can be written as:
\begin{equation}
\begin{aligned}
    \mathcal{J}_{\mathrm{JSD}}^{\beta}
    =
    {}&
    \beta
    \sum_{v \in \mathcal{V}^{T}_{k}}
    \pi_{T,v}
    \log
    \frac{\pi_{M,v}}{\pi_{T,v}}
    +
    (1-\beta)
    \log
    \frac{\pi_{M,x}}{\pi_{S,x}},
\end{aligned}
\end{equation}
where $x \sim \pi_S$, 
$\pi_{M,v} = \beta \pi_{T,v} + (1-\beta)\pi_{S,v}$, 
and $\beta=0.5$ by default.

As shown in Table~\ref{t1}, stand-alone FKL can not achieve effective training when computed over only a small subset of the vocabulary. This is mainly because top-$k$ FKL constitutes a biased approximation of the full-vocabulary FKL objective, and applying this biased divergence to all sampled tokens can introduce increasingly distorted policy gradients. Therefore, FKL is not suitable as the standalone objective for OPD under constrained vocabulary, while there would be great potential to use FKL enhancing OPD objectives like JSD.

\paragraph{Token Filtering and Reward Clipping.}
To mitigate erroneous policy gradients induced by outlier tokens, existing methods mainly adopt two strategies: entropy-based token filtering and reward clipping. In GRPO \cite{shao2024deepseekmath}, training only on the top $20\%$ high-entropy tokens is commonly used to suppress the interference of less informative tokens and accelerate RL convergence. REOPOLD \cite{ko2026scaling} instead applies reward clipping to reduce the influence of erroneous token-level supervision signals. Specifically, it requires a predefined clipping threshold, and rewards exceeding this threshold are clipped to the corresponding upper bound during training.

As shown in Tables~\ref{t1} and~\ref{main}, entropy-aware token selection \cite{wang2026beyond} does not consistently benefit OPD: restricting optimization to high-entropy tokens often degrades performance. This suggests that, in the OPD setting, the teacher can also provide sufficiently informative supervision on ordinary tokens, which should not be discarded during training. Reward clipping improves model performance in Table~\ref{t1}; however, as shown in Table~\ref{main}, its gains become marginal under other settings. Moreover, the choice of clipping threshold introduces an additional hyperparameter that may substantially affect the learning dynamics and convergence speed.

\begin{figure}[t!]
\centering
\begin{subfigure}[t]{0.48\columnwidth}
    \centering
    \includegraphics[width=\linewidth]{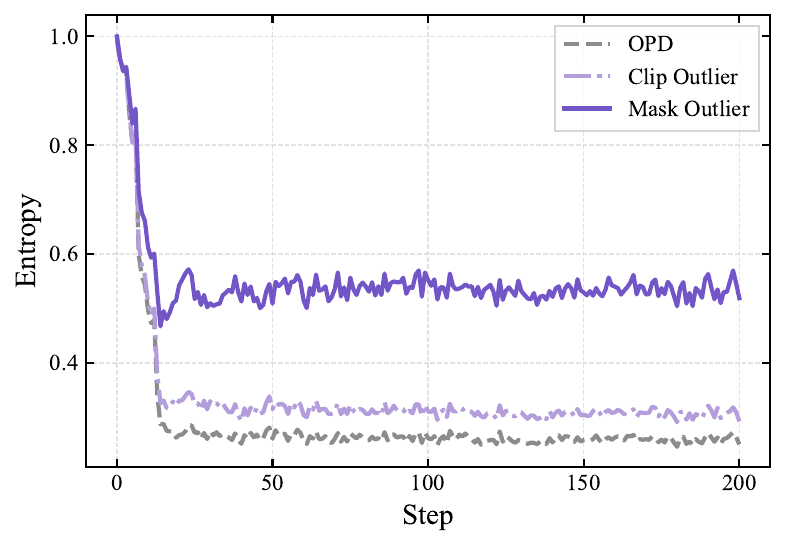}
    \vspace{-0.25in}
    \caption{Entropy comparison.}
    \label{fig:3}
\end{subfigure}
\hfill
\begin{subfigure}[t]{0.48\columnwidth}
    \centering
    \includegraphics[width=\linewidth]{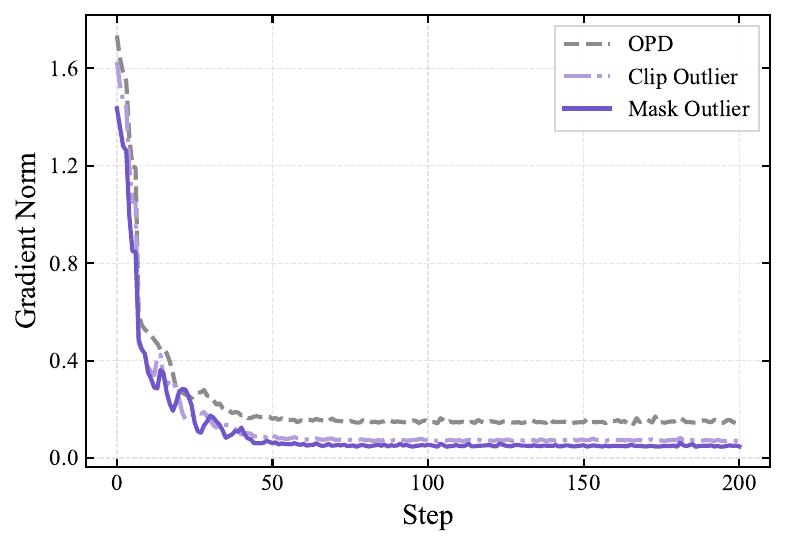}
    \vspace{-0.25in}
    \caption{Gradient norm comparison.}
    \label{fig:4}
\end{subfigure}
\caption{Comparison of OPD methods in terms of entropy and gradient norm.}
\label{fig:entropy_grad_comparison}
\vspace{-0.05in}
\end{figure}

\subsection{Trust-Region On-Policy Learning.}

Based on the benchmark results, directly applying conventional FKL alone fails to achieve effective training, while FKL should be properly combined with RKL to enhance the effectiveness of RKL under large distributional mismatch.
Meanwhile, simple reward clipping and entropy-based token selection provide only limited correction to the policy gradients. We therefore focus on enabling more effective learning while explicitly suppressing unreliable policy-gradient.

Inspired by trust-region policy optimization (TRPO) in RL, we propose Trust Region On-Policy Distillation (TrOPD) that only optimizes where the policy-gradient is reliable. 
Given the trust-region $\color{maskpurple}\mathbb{M}_{x}$ and the outlier $\color{maskpurple}\overline{\mathbb{M}}_{x}$, the token-level objective in $x$ can be governed by:
\begin{equation}
\begin{aligned}
\mathcal{J}_{x}^{On}
={}&
-{\color{maskpurple}\mathbb{M}_{x}}
    \mathrm{KL}(\pi_S \parallel \pi_T)
-{\color{maskpurple}\overline{\mathbb{M}}_{x}}
    {\color{reversemagenta}
    \mathrm{KL}(\pi_T \parallel \pi_S)}
\\
={}&
-{\color{maskpurple}\mathbb{M}_{x}}
    \log \frac{\pi_S}{\pi_T}
-{\color{maskpurple}\overline{\mathbb{M}}_{x}}
    {\color{reversemagenta}
    \sum_{v \in \mathcal{V}^{T}_{k}}
    \pi_{T,v}
    \log \frac{\pi_{T,v}}{\pi_{S,v}}
    } .
\end{aligned}
\end{equation}
Since $x \sim \pi_S$, we estimate RKL within the trust region using the $K_1$ estimator, while approximating the FKL term for outliers using a Top-$k$ estimator:
$
{\color{reversemagenta}
\sum_{v \in \mathcal{V}^{T}_{k}}
\pi_{T,v}
\log \frac{\pi_{T,v}}{\pi_{S,v}}
}$.
In the following, we reason the TrOPD objective step by step.

\paragraph{Outlier Masking.} 
We focus on the first term, 
$-{\color{maskpurple}\mathbb{M}_{x}}\log \frac{\pi_S}{\pi_T}$
, and investigate whether effective training can still be achieved after removing outliers. For a fair comparison, we temporarily adopt a static threshold as \textit{Clip Outlier} in REOPOLD. Different from \textit{Clip Outlier}, which clips rewards exceeding the threshold $\tau$, $\mathrm{R}(x)=\mathrm{max}(\log \frac{\pi_S}{\pi_T},\tau)$, \textit{Mask Outlier} directly masks the token-level advantage once its reward magnitude exceeds the threshold, $\mathrm{R}(x)=\mathbb{I}[\mathrm{R}>\tau]\log \frac{\pi_S}{\pi_T}$.

As shown in Table~\ref{t1}, both \textit{Mask Outlier} and \textit{Clip Outlier} outperform the vanilla OPD baseline, demonstrating that suppressing outlier policy gradients improves optimization stability and downstream performance. 
As shown in Figures~\ref{fig:3} and~\ref{fig:4}, \textit{Mask Outlier} directly eliminates the influence of unreliable gradients. Compared with OPD and \textit{Clip Outlier}, it maintains higher policy entropy, thereby better preserving the exploration capability during OPD training. Moreover, by removing the gradients induced by outlier tokens during backpropagation, TrOPD achieves a lower gradient norm than OPD and \textit{Clip Outlier}, leading to more stable optimization.

\paragraph{Adaptive Trust Region.}
Different from previous predefined threshold $\tau$, we define the  trust region according to student policy $\pi_S(x)$ and teacher check $\pi_T(x)$.
For each token sampled from the student generation, $x \sim \pi_S$, 
the probability of being classified into the trust region, ${\color{maskpurple}\mathbb{M}_{x}}
\sim
\mathrm{Bernoulli}\!\left(P_{\mathrm{trust}}(x)\right)$, is defined as:
\begin{equation}
P_{\mathrm{trust}}(x)
=
\min\left(
\frac{\pi_T(x)}{\pi_S(x)},
1
\right).
\end{equation}
This design is motivated by speculative decoding, where the probability that the teacher model agrees with a token decoded by the student model satisfies $P_{\mathrm{accept}}(x) \propto \min\left(\frac{\pi_T(x)}{\pi_S(x)}, 1\right)$. By selecting only the decoding regions accepted by the teacher, it can be ensured that the student model remains within effectively supervised regions under the $K_1$ estimator.

\paragraph{Outlier Estimation.}
Outlier regions exhibit substantial distributional mismatch between the teacher and student, but may still contain informative supervisory signals. Simply masking these regions may therefore discard useful knowledge. To partially recover such supervision, we introduce an auxiliary forward KL (FKL) objective in outlier regions. Specifically, because reverse KL estimated from student-sampled tokens may fail to provide reliable supervision under severe distributional mismatch, we instead compute the distillation signal from the teacher perspective:
\begin{equation}
\mathcal{J}_{x}^{FKL}
=
-{\color{maskpurple}\overline{\mathbb{M}}_{x}}
    {\color{reversemagenta}
    \sum_{v \in \mathcal{V}^{T}_{k}}
    \pi_{T(v)}
    \log \frac{\pi_{T(v)}}{\pi_{S(v)}}
    },
\end{equation}
where $\mathcal{V}_{T}^{k}=\operatorname{TopK}(\pi_T)$ denotes the teacher's top-$k$ vocabulary.
When $\exists\, v \in \mathcal{V}_{T}^{k} \quad \text{s.t.} \quad \pi_S(v) > 0$, the FKL objective enables imitation learning from informative teacher-supported tokens in the outlier region. In contrast, when
$\sum_{v \in \mathcal{V}_{T}^{k}} \pi_S(v) \rightarrow 0$,
we have $\mathrm{KL}(\pi_T \parallel \pi_S) \rightarrow 0$, such that the auxiliary outlier objective is suppressed and does not interfere with gradient in the trust region. Therefore, this design alleviates the potential loss of supervisory information caused by region masking while preserving stable trust-region optimization.

\begin{table*}[t!]
\centering
\small
\setlength{\tabcolsep}{10pt}
\renewcommand{\arraystretch}{1.18}
\resizebox{\textwidth}{!}{%
\begin{tabular}{lcccc}
\toprule
\textbf{Region} & \textbf{Policy} & \textbf{Objective} & \textbf{Estimator} & \textbf{Memory} \\
\midrule
On-Policy Trust Region 
& $x \sim \pi_S$ 
& $-\mathrm{KL}(\pi_S \| \pi_T)$
& $\displaystyle \log \frac{\pi_T(x)}{\pi_S(x)}$
& $\mathcal{O}(n)$ \\

On-Policy Outlier 
& $x \sim \pi_S$ 
& $-\mathrm{KL}(\pi_T \| \pi_S)$
& $\displaystyle \sum_{v \in \mathcal{V}_{T}^{(k)}} 
\pi_T(v)\log \frac{\pi_S(v)}{\pi_T(v)}$
& $\mathcal{O}(nk)$ \\

Off-Policy Guidance 
& $x \sim \pi_T$ 
& $-\beta \mathrm{KL}(\pi_T \| \pi_S)$
& $\displaystyle \beta \log \frac{\pi_S(x)}{\pi_T(x)}$
& $\mathcal{O}(n)$ \\
\bottomrule
\end{tabular}}
\vskip -0.05in 
\caption{
Region-specific learning objectives and estimators in TrOPD.
Here, $\mathcal{V}_{T}^{(k)}$ denotes the top-$k$ vocabulary under the teacher distribution $\pi_T$.
}
\label{tab:region_objectives}
\vskip -0.1in 
\end{table*}

\subsection{Off-Policy Trust-Region Guidance}

To guide the student model to follow the teacher's trajectory, we propose an off-policy trust region to provide offline constraints, as illustrated in Fig.~\ref{f2}.
The distillation trajectory consists of two parts: an off-policy prefix $x[:l]$ generated by the teacher, followed by an on-policy continuation $x[l:]$ generated by the student.
For complex reasoning tasks, this design avoids low-quality outputs caused by the limited capability of the student model.
We apply forward KL, $\mathrm{KL}_{x[:l] \sim \pi_T}(\pi_T \parallel \pi_S)$, for imitation learning from the off-policy guidance:
\begin{equation}
\begin{aligned}
\mathcal{J}_x
&=
-
\textcolor{deepblue}{\beta\mathrm{KL}_{x[:l] \sim \pi_T}(\pi_T \parallel \pi_S)}
+
\mathcal{J}^{\mathrm{On}}_{x[l:]} \\
&=
-
\textcolor{deepblue}{\beta \mathbb{I}[x \sim \pi_T]
\log \frac{\pi_T}{\pi_S}}
+
\textcolor{deepblue}{\mathbb{I}[x \sim \pi_S]}
\mathcal{J}^{\mathrm{On}}_{x[l:]}
\end{aligned}
\end{equation}
For the off-policy region, since samples are generated from the teacher, $x \sim \pi_T$, we adopt the $K_1$ estimator for the forward KL, $\mathrm{KL}(\pi_T \parallel \pi_S) = \log \frac{\pi_S}{\pi_T}$, which achieves $\mathcal{O}(n)$ memory complexity.

\paragraph{Unified Optimization.}
We summarize the overall objective of TrOPD as:

\begin{equation}
\begin{aligned}
\mathcal{J}_{x}^{\mathrm{TrOPD}}
&=
-{
  \mathbb{I}[x \sim \pi_S]\overline{\mathbb{M}}_{x}}
 {
  \sum_{v \in \mathcal{V}^{T}_{k}}
  \pi_{T,v}
  \log \frac{\pi_{T,v}}{\pi_{S,v}}
 } \\
&\quad
-
{
  \mathbb{I}[x \sim \pi_S]\mathbb{M}_{x}}
\log \frac{\pi_S}{\pi_T}
-
\beta \,
\mathbb{I}[x \sim \pi_T]
\log \frac{\pi_T}{\pi_S} .
\end{aligned}
\end{equation}

where the details of each component are in Table~\ref{tab:region_objectives}. Initially, the maximum off-policy trajectory length is set to the maximum training sequence length. During training, it is gradually annealed to zero using a cosine schedule, such that generation becomes fully on-policy by the end of training.

\section{Experimental Results}

\begin{table*}[t]
\centering
\small
\setlength{\tabcolsep}{6pt}
\renewcommand{\arraystretch}{1.1}
\resizebox{\textwidth}{!}{%
\begin{tabular}{lcccccc}
\toprule
Method & AIME 24 & AIME 25 & AMC 23 & LiveCodeBench v6 & GPQA diamond & Avg. \\
\midrule
DeepSeek-Qwen2.5-1.5B & 28.64 & 24.16 & 71.01 & 15.43 & 34.22 & 34.69 \\
\midrule

\multicolumn{6}{l}{\emph{\quad \textbf{Single-Domain Distillation}}} \\ 
Teacher & 66.14 & 51.87& 92.34 & 34.86 & 47.22 & 58.48 \\
OPD                & 35.83 & 29.16 & 75.39 & 17.14 & 28.03 & 37.11 \\
EOPD               & 36.97 & 29.79 & 75.23 & 15.43 & 32.58 & 38.00 \\
Entropy OPD 20\%   & 35.52 & 29.06 & 73.82 & 14.29 & 31.82 & 36.90 \\
REOPOLD 2Stage & 34.47 & 29.89 & 73.35 & 16.57 & 30.18 & 36.89 \\
REOPOLD        & 36.97 & 30.83 & 75.78 & 18.29 & 32.07 & 38.79 \\
\rowcolor{lightpurple} TrOPD 
                   & 38.54 & 32.50 & 77.03 & 18.86 & 36.24 & 40.63 \\
\midrule

\multicolumn{6}{l}{\emph{\quad \textbf{Multi-Domain Distillation}}} \\ 
Teacher & 65.62  & 52.81 & 91.79 & 36.57 & 47.22 & 58.80 \\
OPD                          & 30.10 & 21.66 & 61.56 & 20.57 & 31.06 & 32.99 \\
REOPOLD                  & 34.27 & 25.83 & 63.90 & 19.43 & 34.47 & 35.58 \\
\rowcolor{lightpurple} TrOPD & 36.04 & 27.60 & 70.93 & 22.29 & 31.19 & 37.61 \\
\bottomrule
\end{tabular}}
\caption{Performance comparison using DeepSeek-R1-Distill-Qwen-1.5B as the student model. Skywork-OR1-Math-7B and Skywork-OR1-7B are teacher models for the single-domain and multi-domain distillation respectively.}
\label{tab:opd_results}
\end{table*}

\begin{table*}[t]
\centering
\small
\setlength{\tabcolsep}{6pt}
\renewcommand{\arraystretch}{1.1}
\resizebox{\textwidth}{!}{
\begin{tabular}{lcccccccc}
\toprule
\multirow{2}{*}{\textbf{Method}} 
& \multicolumn{3}{c}{\textbf{Math}} 
& \multicolumn{2}{c}{\textbf{STEM}} 
& \multicolumn{1}{c}{\textbf{Instruct}} 
& \multicolumn{1}{c}{\textbf{Code}} 
& \multirow{2}{*}{\textbf{Avg.}} \\
\cmidrule(lr){2-4}
\cmidrule(lr){5-6}
\cmidrule(lr){7-7}
\cmidrule(lr){8-8}
& AIME 24 & AIME 25 & AMC 23 & GPQA dia. & MMLU red. & IFBench & LCB.v6 & \\
\midrule
Qwen3-SFT-1.7B & 35.41 & 26.45 & 68.90 & 25.25 & 66.60 & 26.19 & 30.29 & 39.87 \\
\midrule
\multicolumn{6}{l}{\emph{\quad \textbf{Multi-Domain Distillation}}} \\ 
Teacher            & 81.66 & 75.72 & 98.98 & 58.86 & 77.03 & 62.93 & 58.86 & 73.43 \\
OPD                & 48.02 & 40.72 & 81.79 & 29.80 & 68.60 & 37.07 & 32.00 & 48.29 \\
EOPD               & 47.08 & 40.83 & 81.32 & 33.84 & 68.26 & 36.39 & 34.29 & 48.86 \\
Entropy OPD        & 43.54 & 42.70 & 79.53 & 29.92 & 68.51 & 38.78 & 33.71 & 48.10 \\
REOPOLD        & 45.62 & 42.29 & 81.64 & 30.56 & 68.30 & 36.05 & 35.43 & 48.56 \\
\rowcolor{lightpurple} TrOPD
                   & 52.08 & 44.06 & 83.04 & 35.98 & 68.74 & 42.18 & 36.00 & 51.73 \\
\bottomrule
\end{tabular}}
\caption{Performance comparison using Qwen3-SFT-1.7B as the student model. Qwen3-Nemotron-4B is the teacher model for the multi-domain distillation.}
\label{main}
\vskip -0.1in 
\end{table*}

\subsection{Implementation Details}

Without loss of generality, we benchmark OPD for reasoning models in two settings, i.e., single-domain and multi-domain reasoning distillation.

\paragraph{Model.}
(1) For single-task distillation with DeepSeek-Distilled-Qwen-1.5B \cite{zhang2025100}, we use the representative mathematical reasoning task, where Skywork-OR1-Math-7B \cite{he2025skywork} serves as the teacher model and DeepSeek-Distilled-Qwen-1.5B serves as the student model. Since the student has already undergone extensive off-policy distillation, we directly conduct OPD without additional SFT.
(2) For multi-task distillation with DeepSeek-Distilled-Qwen-1.5B, we use Skywork-OR1-7B \cite{he2025skywork} as the teacher model and DeepSeek-Distilled-Qwen-1.5B as the student model to evaluate OPD in multi-domain reasoning tasks.
(3) For multi-task distillation with Qwen3-SFT-1.7B \cite{wang2026mix}, the teacher model, Qwen3-Nemotron-4B \cite{wang2026mix}, is trained from Qwen3-4B-Base \cite{yang2025qwen3} using Nemotron SFT \& RL data and recipe (Appendix A for details). The student model is trained from Qwen3-1.7B-Base using the same SFT recipe.

\paragraph{Dataset.}
(1) For single-domain distillation, we use prompts from the OpenThoughts3 dataset \cite{guha2025openthoughts} and retain only examples from the mathematics domain.
(2) For multi-domain distillation, we use prompts from OpenThoughts3, covering the mathematics, code, and science domains. Only the prompts are retained for training.

\begin{table}[t]
\centering
\small
\setlength{\tabcolsep}{6pt}
\renewcommand{\arraystretch}{1.1}
\begin{tabular}{lccccc}
\toprule
Method & Outlier Objective & AIME 24 & AIME 25 & AMC 23 & Avg. \\
\midrule
\multicolumn{3}{l}{\emph{\quad \textbf{Single-Domain Distillation}}} \\
DS-1.5B            & -- & 28.64 & 24.16 & 71.01 & 41.27 \\
OPD                & $\log \pi_{T}/\pi_{S}$
                   & 35.83 & 29.16 & 75.39 & 46.79 \\
\midrule
\multicolumn{3}{l}{\emph{\quad \textbf{+ Outlier Estimation}}} \\ 
Mask Outlier       & 0  & 37.08 & 30.62 & 75.46 & 47.72 \\
Clip Outlier       & $\tau$ & 36.97 & 30.83 & 75.78 & 47.86 \\
Full FKL           & $\sum_{v \in \mathcal{V}^{T}_{k}} \pi_{T,v} \log (\pi_{S,v}/\pi_{T,v})$ 
                   & 0.00  & 0.00  & 4.21  & 1.40  \\
\rowcolor{lightpurple} FKL Outlier 
                   & $\sum_{v \in \mathcal{V}^{T}_{k}} \pi_{T,v} \log (\pi_{S,v}/\pi_{T,v})$ 
                   & 39.16 & 29.89 & 77.96 & 49.00 \\
\midrule
\multicolumn{3}{l}{\emph{\quad \textbf{+ Off-Policy Guidance}}} \\ 
TrOPD Mask         & 0 & 40.10 & 30.41 & 75.85 & 48.79 \\
TrOPD Clip         & $\tau$ & 37.39 & 31.77 & 77.03 & 48.73 \\
\rowcolor{lightpurple} TrOPD FKL 
                   & $\sum_{v \in \mathcal{V}^{T}_{k}} \pi_{T,v} \log (\pi_{S,v}/\pi_{T,v})$ 
                   & 38.54 & 32.50 & 78.51 & 49.85 \\
\bottomrule
\end{tabular}
\caption{Ablation Studies of TrOPD in math-domain distillation.}
\vskip -0.1in 
\label{tab:ablation}
\end{table}

\begin{table*}[!t]
\centering
\small
\setlength{\tabcolsep}{6pt}
\renewcommand{\arraystretch}{1.1}
\resizebox{\textwidth}{!}{%
\begin{tabular}{lcccccc}
\toprule
Method & AIME 24 & AIME 25 & AMC 23 & LiveCodeBench v6 & GPQA diamond & Avg. \\
\midrule
DeepSeek-Qwen2.5-1.5B        & 28.64 & 24.16 & 71.01 & 15.43 & 34.22 & 34.69 \\
OPD                           & 35.83 & 29.16 & 75.39 & 17.14 & 28.03 & 37.11 \\
\midrule
AOPD                          & 39.89 & 30.00 & 77.18 & 20.57 & 31.31 & 39.79 \\
\rowcolor{lightpurple} TrOPD  & 38.54 & 32.50 & 77.03 & 18.86 & 36.24 & 40.63 \\
\rowcolor{lightpurple} TrOPD + AOPD 
                              & 42.08 & 31.87 & 78.20 & 21.71 & 34.47 & 41.67 \\
\bottomrule
\end{tabular}}
\caption{Performance comparison between TrOPD and concurrent AOPD. TrOPD + AOPD indicates TrOPD adding the AOPD objective for the AOPD positive samples.}
\label{tab:aopd_comparison}
\end{table*}

\paragraph{Benchmark Training.}
To fairly compare existing OPD methods, we train all benchmarks using the same training settings. Specifically, we perform OPD training for 200 steps using a fixed learning rate of $5 \times 10^{-6}$. For FKL-based methods implemented with a top-$k$ support set, we uniformly set $k=64$. For off-policy guidance, we set $\beta=0.001$ for imitation learning. We use a prompt batch size of 128 and sample 4 rollouts for each prompt, with a maximum generation length of 8096 tokens.

\paragraph{Benchmark Evaluation.}
We evaluate the distilled models across mathematics, STEM, and instruction following domains. For mathematical reasoning, we report results on AIME 2024, AIME 2025 \cite{shi2025aime}, and AMC 2023, while each result is the average accuracy of 32 times evaluation. For STEM reasoning and instruction following, we use GPQA Diamond, MMLU-Redux v2 \cite{gema2025we}, and IFBench \cite{pyatkin2025generalizing}. For code generation, we evaluate on LiveCodeBench v6 \cite{jain2025livecodebench}.

\subsection{Main Results}

\paragraph{Single-Domain Distillation.}
As shown in Table~\ref{tab:opd_results}, we primarily evaluate mathematical reasoning performance on AIME 2024, AIME 2025, and AMC 2023. We further report results on out-of-domain (OOD) tasks to assess the continual learning capability of OPD methods and their robustness to domain shifts. Compared with OPD, TrOPD improves the average performance by +3.06 points on mathematical reasoning tasks and by +2.63 points on general-domain tasks. REOPOLD corrects unreliable policy gradients using simple reward clipping; nevertheless, TrOPD outperforms it by 1.99 and 1.84 points in the mathematical and general domains, respectively, demonstrating the necessity of trust-region learning and outlier estimation. Furthermore, compared with entropy-based token selection methods, including EOPD \cite{jin2026entropy}, Entropy OPD \cite{ko2026scaling, wang2026beyond}, and REOPOLD 2Stage \cite{ko2026scaling}, TrOPD achieves improvements of 2.63, 3.73, and 3.74 points, respectively. These results indicate that outlier-aware token selection provides a more effective criterion than entropy-based selection.

\paragraph{Multi-Domain Distillation.}
As shown in Table~\ref{tab:opd_results} and Table~\ref{main}, we evaluate multi-domain distillation performance with Skywork-OR1-Math-7B and Qwen3-Nemotron-4B as the teacher models, respectively. Since Skywork-OR1-Math-7B is primarily trained for mathematical reasoning and code generation, we mainly evaluate its distilled domains on AIME 2024, AIME 2025, AMC 2023, and LiveCodeBench. Compared with OPD, TrOPD consistently improves the performance of both DeepSeek-Qwen2.5-1.5B and Qwen3-SFT-1.7B, achieving substantial average gains of +4.62 and +3.44 points, respectively. These results demonstrate that TrOPD can consistently improve distillation performance across different teacher--student configurations and diverse reasoning tasks.

\subsection{Ablation Studies and Discussion}

As shown in Table~\ref{tab:ablation}, applying FKL only to outlier regions achieves better performance than masking or clipping outliers, demonstrating its effectiveness for outlier estimation. Incorporating off-policy guidance further improves the average scores of the Clip, Mask, and FKL Outlier variants, confirming its complementary benefit. Consequently, the three TrOPD variants, i.e., TrOPD Mask, TrOPD Clip, and TrOPD FKL, outperform OPD by 2.00, 1.94, and 3.06 points on average, respectively.

We also notice the concurrent work AOPD \cite{jia2026asymmetric}. As shown in Table~\ref{tab:aopd_comparison}, TrOPD outperforms AOPD, while their combination further improves the average score from 40.63 to 41.67. This suggests that AOPD is orthogonal to TrOPD, and combining complementary OPD optimization strategies is a promising direction for future work.

\section{Conclusion}
This work proposes Trust Region On-Policy Distillation, a reliable and stable framework for reasoning-oriented OPD. By trust region optimization and outlier estimation, TrOPD effectively suppresses unreliable policy gradients while preserving informative supervision. 
We further introduce off-policy guidance to encourage exploration toward teacher-supported trajectories. 
Extensive multi-domain results highlight the importance of supervision reliability in on-policy reasoning distillation and demonstrate the future potential of trust-region learning for training high-quality small reasoning models.

\section*{Limitations}

The primary limitation of this work is the lack of practical deployment and application studies on small reasoning models. In real-world scenarios, training high-performing small reasoning models often requires incorporating mid-training to further improve their post-training reasoning capabilities. This work focuses primarily on OPD-based post-training using DeepSeek-Qwen2.5-1.5B and Qwen3-SFT-1.7B, which may constrain the upper bound of the resulting reasoning performance. Future work should investigate how additional stages, such as pre-training and mid-training, can further enhance small reasoning models in practical deployment settings. Nevertheless, this study focuses on fair and controlled comparisons among OPD methods.

\bibliography{custom}
\bibliographystyle{colm2026_conference}

\newpage
\appendix
\onecolumn

\section*{Appendix}
\section{Training Details of Teacher Model Qwen3-Nemotron-4B}\label{sec:appendix}

The training pipeline is initialized from Qwen3-4B-Base and consists of both supervised fine-tuning (SFT) and reinforcement learning with verifiable rewards (RLVR). For SFT, publicly available datasets released with Nemotron 3 Nano\footnote{\url{https://huggingface.co/collections/nvidia/nemotron-post-training-v3}}~\citep{blakeman2025nemotron} are adopted. Entries without the \texttt{messages} field are removed, and datasets from multiple domains are combined. To match the domain distribution ratios reported in the corresponding technical report, smaller datasets are upsampled while larger datasets are randomly downsampled, resulting in approximately 14M training samples.

For RLVR, the publicly available training blend released with Nemotron 3 Nano\footnote{\url{https://huggingface.co/datasets/nvidia/Nemotron-3-Nano-RL-Training-Blend}} is adopted. The resulting training mixture covers four domains: (1) \textit{Math}: 22,056 samples from DAPO~\citep{yu2025dapo} and Skywork~\citep{he2025skywork,skywork-or1-2025}; (2) \textit{Coding}: 19,169 samples from CodeContests~\citep{li2022competition} and Open-R1~\citep{penedo2025codeforces}; (3) \textit{Science}: 19,670 samples from OpenScienceReasoning-2~\citep{open_science_reasoning_2_2025}; and (4) \textit{Instruction Following}: 16,575 samples from WildChat-1M~\citep{zhao2024wildchat}, with instructions sourced from Open-Instruct~\citep{lambert2024tulu}.

During SFT, the Adam optimizer~\citep{kingma2014adam} is used with a learning rate of $5 \times 10^{-5}$ and a weight decay of $0.1$. The warmup phase accounts for $10\%$ of the total training steps. The batch size is set to 512, with an average response length of approximately 7K tokens.

During RLVR, GRPO is employed with a group size of 16, together with masked importance sampling to improve consistency between training and inference. The batch size is set to 128, and model parameters are updated every 2048 rollouts. The maximum generation length is capped at 32K tokens, and the sampling temperature is set to 1.0 to encourage exploration. 
More details on the training recipe of the multi-task reinforcement learning can be found in Wang et al.~\cite{wang2026mix}.

\end{document}